\documentclass[journal]{IEEEtran}
\usepackage{cite}
\usepackage{amsmath,amssymb,amsfonts}
\usepackage{algorithmic}
\usepackage{graphicx}
\usepackage{subcaption}
\usepackage{textcomp}
\usepackage{multirow}
\usepackage{booktabs}
\usepackage{siunitx}
\usepackage[breaklinks=true,bookmarks=false]{hyperref}

\begin{document}
\title{FlatteNet: A Simple Versatile Framework for Dense Pixelwise Prediction}
\author{Xin~Cai, Yi-Fei~Pu\thanks{This work was supported in part by the National Natural Science Foundation of China under Grant 61571312 and in part by the National Key Research and Development Program Foundation of China under Grant 2018YFC0830300 and Grant 2017YFB0802300. \textit(Corresponding author: Yi-Fei~Pu.)}
\thanks{X. Cai is with the College of Computer Science \& Software Engineering, Sichuan University, Chengdu 610065, China (e-mail: xincai00@gmail.com).}
\thanks{Y.-F. Pu is with the College of Computer Science \& Software Engineering, Sichuan University, Chengdu 610065, China, and also with the Research and Development Department, Chengdu PU Chip Science and Technology Company, Ltd., Chengdu 610066, China (e-mail: puyifei@scu.edu.cn).}}

\markboth
{X. CAI \MakeLowercase{\textit{et al.}}: FlatteNet: A Simple Versatile Framework for Dense Pixelwise Prediction}
{X. CAI \MakeLowercase{\textit{et al.}}: FlatteNet: A Simple Versatile Framework for Dense Pixelwise Prediction}

\maketitle

\begin{abstract}
In this paper, we focus on devising a versatile framework for dense pixelwise prediction whose goal is to assign a discrete or continuous label to each pixel for an image. It is well-known that the reduced feature resolution due to repeated subsampling operations poses a serious challenge to Fully Convolutional Network (FCN) based models. In contrast to the commonly-used strategies, such as dilated convolution and encoder-decoder structure, we introduce the Flattening Module to produce high-resolution predictions without either removing any subsampling operations or building a complicated decoder module. In addition, the Flattening Module is lightweight and can be easily combined with any existing FCNs, allowing the model builder to trade off among model size, computational cost and accuracy by simply choosing different backbone networks. We empirically demonstrate the effectiveness of the proposed Flattening Module through competitive results in human pose estimation on MPII, semantic segmentation on PASCAL-Context and object detection on PASCAL VOC. We hope that the proposed approach can serve as a simple and strong alternative of current dominant dense pixelwise prediction frameworks.
\end{abstract}


\section{Introduction}
\label{sec:introduction}
\IEEEPARstart{M}{any} fundamental computer vision tasks can be formulated as a dense pixelwise prediction problem. Examples include but are not limited to: semantic segmentation \cite{chen2018encoder}, human pose estimation \cite{xiao2018simple}, saliency detection \cite{zhang2018progressive}, depth estimation \cite{silberman2012indoor}, optical flow \cite{ilg2017flownet}, super-resolution \cite{shi2016real} and image generation with Generative Adversarial Networks (GANs) \cite{isola2017image}. In addition, there is a growing interest in reducing anchor boxes based object detection to a pixelwise prediction problem \cite{law2018cornernet}, \cite{tian2019fcos}, \cite{zhou2019objects}, \cite{duan2019centernet}. It is therefore desirable to devise a versatile framework that can effectively and efficiently tackle the dense pixelwise prediction problem.

\begin{figure*}[t!]
\centering\includegraphics[scale=1.6]{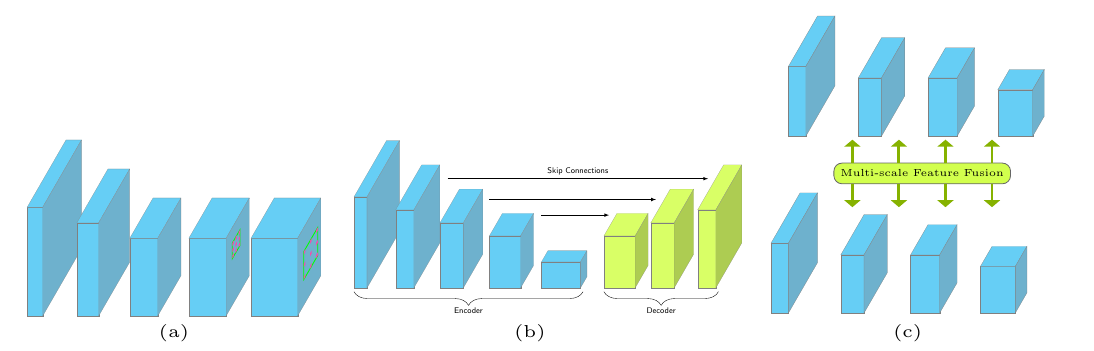}
\caption{An illustration of existing representative networks that are employed to tackle challenging dense prediction tasks: (a) denotes the dilated FCN; (b) denotes the encoder-decoder network; (c) denotes the multi-stream network.}
\label{fig:fig1}
\end{figure*}

Recently, deep learning methods, and in particular deep convolutional neural networks (DCNNs) based on the Fully Convolutional Network (FCN) framework \cite{long2015fully},  have achieved tremendous success in such dense pixelwise prediction tasks. However, it is well-known that the major issue for current FCN based models is the reduced feature resolution caused by the repeated combination of spatial pooling and convolution striding performed at consecutive layers of DCNNs which have been originally designed for image classification \cite{krizhevsky2012imagenet}, \cite{simonyan2014very}, \cite{he2016deep}. Various techniques have been proposed in order to overcome this limitation and generate high-resolution feature maps. As illustrated in Fig.\ref{fig:fig1}, we mainly consider three categories in this work. (a) Dilated convolution is used to repurpose ImageNet \cite{deng2009imagenet} pre-trained networks to extract denser feature maps by removing the subsampling operations from the last few layers, e.g., \cite{chen2018encoder}, \cite{chen2017rethinking}, \cite{chen2017deeplab}. The major drawback of dilated convolution based networks is computationally prohibitive and demanding large GPU memory due to the processing of high-dimensional and high-resolution feature maps. (b) Many state-of-the-art dense pixelwise prediction models belong to the family of encoder-decoder networks, e.g., \cite{xiao2018simple}, \cite{ronneberger2015u}, \cite{newell2016stacked}, \cite{lin2017refinenet}. First, an encoder sub-network with subsampling operations decreases the spatial resolution (usually by a factor of $32$) while increasing the number of channels. Afterwards, a decoder sub-network upsamples the low-resolution feature maps back to the original input resolution. In spite of their impressive performance, the network architectures have become increasingly complex, especially the decoder modules, which results in much more parameters as well as computational complexity. Besides, the high complexity is adverse to clear idea validation and fair experimental comparison. (c) The other networks are composed of multiple parallel streams from the input image to the output prediction, working at different spatial resolutions \cite{fourure2017gridnet}, \cite{sun2019high}, \cite{saxena2016convolutional}. High resolution streams allow the network to give accurate predictions in combination with low resolution streams which carry strong semantics. The clear downside of such methods is that they cannot re-use a wide range of pre-trained image classification networks that are readily available for the community, and thus requiring expensive training from scratch.

Despite the approaches mentioned above having made great progress, it remains an open question how to generate high-resolution predictions in an efficient and elegant way. The central premise of dilated convolution based models is that the subsampling operations are detrimental to dense prediction tasks where high-resolution predictions are expected. This work starts from questioning this premise: \textit{Is it truly necessary to sacrifice the benefits of subsampling operations, such as effectively increasing the receptive field size and reducing computational complexity, for spatial prediction accuracy?} In addition, the impressive performance of Simple~Baseline \cite{xiao2018simple}, which simply adds a few deconvolutional layers on top of a backbone network, leads us to the second question: \textit{Is it truly necessary to build a sophisticated decoder to attain solid performance?} These two preliminary questions motivate us to reconsider the current paradigms of solving the dense pixelwise prediction problem by exploring the possibility of making accurate dense pixelwise predictions directly using the coarse-grained features outputted by a FCN, resulting in our quite simple yet surprisingly effective approach.

Our main contributions can be summarized as follows:
\begin{itemize}
\item[•] We introduce a novel scheme to produce dense pixelwise predictions based on the proposed lightweight Flattening Module, while avoiding either removing any subsampling operations or building a complex decoder module. A FCN equipped with the Flattening Module, which we refer to as FlatteNet, can accomplish various dense prediction tasks in an effective and efficient manner. Furthermore, we offer a fresh viewpoint of the proposed Flattening Module to highlight its simplicity.
\item[•] To demonstrate the effectiveness of the proposed Flattening Module, we conduct extensive experiments on three distinct and highly competitive benchmark tasks: MPII Human Pose Estimation task \cite{andriluka2014}, PASCAL-Context Semantic Segmentation task \cite{mottaghi2014}, PASCAL VOC Object Detection task \cite{everingham2015}. Compared to its decoder based or dilation convolution based counterpart, FlatteNet achieves comparable or even better accuracy with much fewer parameters and FLOPs (i.e. the number of floating-point multiplication-adds). 
\end{itemize}

The rest of this paper is organized as follows. In Section \ref{sec:relatedwork}, we briefly review three major strategies which have been developed for dense prediction tasks. Section \ref{sec:flattenet} presents a detailed description of the proposed approach. In section \ref{sec:experimentsandanalysis}, we carry out comprehensive experiments on three challenging benchmark datasets, providing with implementation details and experimental results. Section \ref{conclusionandfutureworks} presents concluding remarks and sketches possible directions for future work.

\section{Related Work}
\label{sec:relatedwork}
State-of-the-art dense pixelwise prediction networks are typically based on the FCN. For conquering the problem of spatial resolution loss caused by subsampling, several methods have been proposed and we mainly consider three categories: dilated convolution \cite{chen2018encoder}, \cite{chen2017rethinking}, \cite{chen2017deeplab}, encoder-decoder architectures \cite{xiao2018simple}, \cite{ronneberger2015u}, \cite{newell2016stacked}, \cite{lin2017refinenet}, \cite{noh2015learning}, and multi-stream networks \cite{fourure2017gridnet}, \cite{sun2019high}, \cite{saxena2016convolutional}.

\subsection{Dilated Convolution}
Dilated (or atrous) convolution is employed to extract denser feature maps by replacing some strided convolutions and associated regular convolutions in classification networks \cite{chen2018encoder}, \cite{chen2017rethinking}, \cite{chen2017deeplab}. With dilated convolution, one is able to control the resolution at which feature responses are computed within DCNNs without requiring learning extra parameters. However, the limitation of dilated convolution based networks is that they need to perform convolutions on a large number of high-resolution feature maps that usually have high-dimensional features, which are computationally prohibitive. Moreover, the processing of a large number of high-dimensional and high-resolution feature maps also require consuming huge GPU memory resources, especially in the training stage.

\subsection{Encoder-Decoder}
The encoder-decoder networks have been successfully applied to many dense prediction tasks. Typically, the encoder-decoder network contains (1) an encoder module that gradually reduces the resolution of feature maps while learning high semantic information, and (2) a decoder module where spatial dimension are gradually recovered. Representative network design patterns fall into two main categories: (1) Symmetric conv-deconv architectures \cite{ronneberger2015u}, \cite{newell2016stacked}, \cite{lin2017refinenet}, \cite{noh2015learning}, \cite{yang2017learning} design the upsampling/deconvolution process as a mirror of the downsampling process and apply skip connections to exploit the features with finer scales. (2) Heavy downsampling process and light upsampling process. The downsampling process is based on the ImageNet \cite{deng2009imagenet} classification network, e.g., ResNet \cite{he2016deep} adopted in \cite{xiao2018simple}, \cite{chen2018cascaded}, and the upsampling process is simply a few bilinear upsampling \cite{chen2018cascaded} or deconvolutional \cite{xiao2018simple} layers. These approaches tend to have much more parameters as well as computational complexity. In addition, the increasingly complex network structures and associated design choices make ablation study and fair comparison difficult.

\begin{figure*}[t!]
\centering\includegraphics[width=0.95\textwidth]{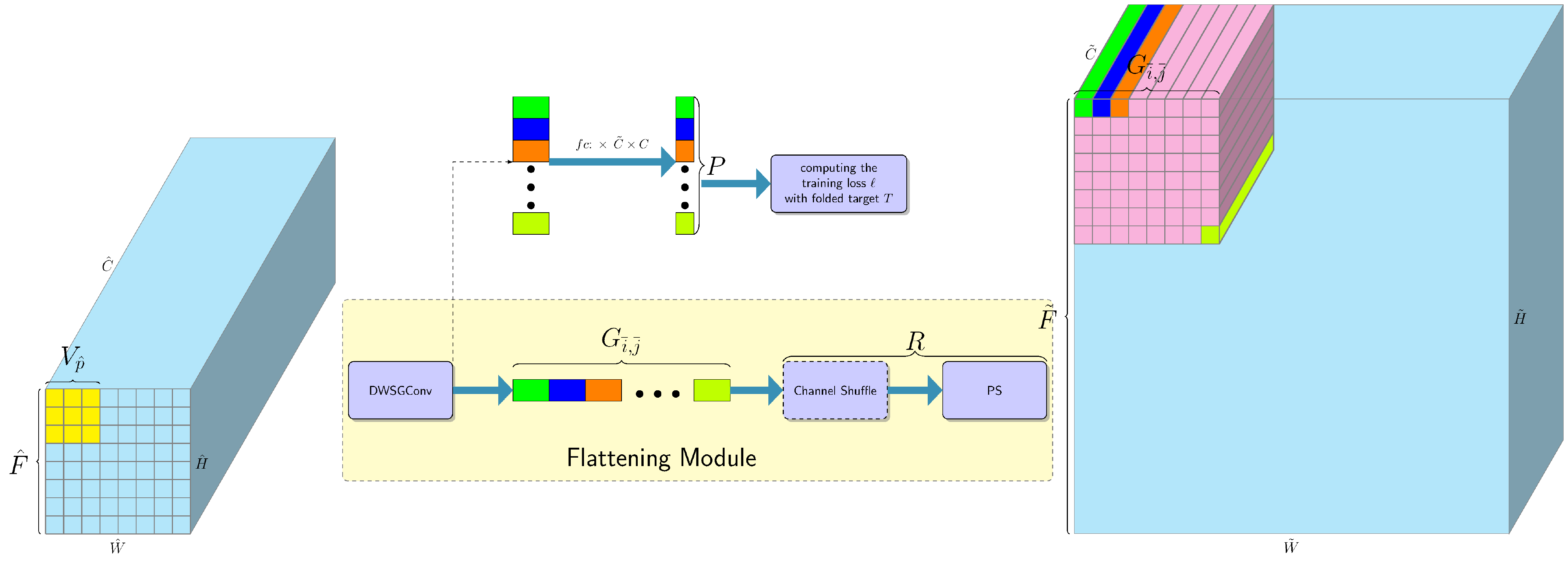}
\caption{An illustration of the proposed Flattening Module. Please refer to the main body for details.}
\label{fig:fig2}
\end{figure*}

\subsection{Multi-stream Network}
This model contains multiple parallel streams from the input image to the output prediction, operating at different spatial resolutions. These streams are interconnected to encode semantic information from multiple scales. Representative works include GridNet \cite{fourure2017gridnet}, convolutional neural fabrics \cite{saxena2016convolutional}, and recently-developed HRNet \cite{sun2019high}. The limitation of these methods is that they cannot utilize a large number of pre-trained models and then fine-tune on target tasks, thus requiring expensive training from scratch.

In contrast to the works mentioned above, our approach generates high-resolution predictions without either removing any subsampling operations or building a complex decoder module, thus significantly reducing the number of parameters and the computational complexity compared to its dilated convolution based or decoder based counterpart. Besides, the proposed Flattening Module can be seamlessly integrated into the FCN framework, thus being able to leverage a large amount of pre-trained classification networks.

\section{FlatteNet}
\label{sec:flattenet}
In this section, we firstly present a general framework for addressing the dense pixelwise prediction problem, from which our specific instantiation is derived, and then introduce the Flattening Module. Finally we offer a different perspective to highlight the simplicity of our method.

\subsection{General Framework for Dense Prediction}
For each pixel location $p$ in a given input image $X\in\mathbb{R}^{3\times H\times W}$, the goal of dense prediction is to compute a discrete label $Y_p\in\lbrace1,\ldots,K\rbrace$ or a continuous label $Y_p\in\mathbb{R}^N$, where $p=\left(i, j\right)$, $i=0,\ldots,H-1$, $j=0,\ldots,W-1$. A general framework modeling the dense prediction problem can be abstracted into three procedures: $\left(a\right)$ learning a set of pixelwise visual descriptors $v_{\tilde{p}}=f_{\tilde{p}}\left(X\right)$, where $\tilde{p}=\left(\tilde{i}, \tilde{j}\right)$, $\tilde{i}=\lfloor\frac{i}{s_1}\rfloor$, $\tilde{j}=\lfloor\frac{j}{s_1}\rfloor$, and $s_1$\footnote{Since natural images exhibit strong spatial correlation, it is not necessary to produce full-resolution predictions. $s_1$ is typically $4$ for encoder-decoder models or $8$ for dilated convolution based models.} is a subsampling factor; $\left(b\right)$ computing the outputs via a pixelwise predictor $Y_{\tilde{p}}=g_{\tilde{p}}\left(v_{\tilde{p}}\right)$; $\left(c\right)$ upsampling the predictions back to the input resolution $Y_p=upsample\left(Y_{\tilde{p}}\right)$. In this work, we mainly consider spatially-invariant feature extractors (e.g., FCNs) and predictors. Therefore, in the following paper, we omit the subscript $\tilde{p}$ for $f$ and $g$.

Further, we can divide the procedure $\left(a\right)$ of learning a set of pixelwise visual descriptors into two sub-procedures: $\left(a_1\right)$ learning a set of patchwise visual descriptors $v_{\hat{p}}=f_1\left(X\right)$, where $\hat{p}=\left(\hat{i}, \hat{j}\right)$, $\hat{i}=\lfloor\frac{\tilde{i}}{s_2}\rfloor$, $\hat{j}=\lfloor\frac{\tilde{j}}{s_2}\rfloor$, $s_2$\footnote{$s_2$ is typically $8$, provided that the resolution of output feature maps of a FCN is $\frac{1}{32}$ of the input image.} is another subsampling factor; $\left(a_2\right)$ generating pixelwise visual descriptors based on these patchwise visual descriptors $v_{\tilde{p}}=f_2\left(V_{\hat{p}}\right)$, where $V_{\hat{p}}\subset\Omega$ is output dependent and $\Omega$ is the set of all patchwise visual descriptors obtained from $\left(a_1\right)$. For example, in the encoder-decoder model, $f_1$, $f_2$, and $g$ correspond to the encoder part, the decoder part, and a simple linear predictor, respectively. 

Many efforts have been devoted to develop a powerful as well as complicated $f_2$ to obtain stronger pixelwise visual descriptors $v_{\tilde{p}}$. Instead, we argue that patchwise visual descriptors $v_{\hat{p}}$, produced by a classification network deployed in a fully convolutional fashion, already contain sufficient information to make accurate pixelwise predictions $Y_{\tilde{p}}$. In other words, a simple $f_2$ that connects these coarse-grained patchwise visual descriptors back to the pixels would perform considerably well.

With the notation introduced above, we can readily formulate our overall framework as follows:
{
\begin{align}
\hat{F}&=f_{\theta_1}\left(X\right)\label{eq:eq1}\\
\tilde{F}&=f_{\theta_2}\left(\hat{F}\right)\label{eq:eq2}\\
Y_{\tilde{p}}&=affine_{\theta_3}\left(v_{\tilde{p}}\right)\label{eq:eq3}\\
Y_p&=bilinear\left(Y_{\tilde{p}}\right)\label{eq:eq4}
\end{align}
}%
where $f_{\theta_1}$ is implemented as a DCNN, e.g., ResNet \cite{he2016deep}, $\hat{F}\in\mathbb{R}^{\hat{C}\times \hat{H}\times \hat{W}}$\footnote{$\hat{C}$ denotes the number of channels of $v_{\hat{p}}$. $\frac{\hat{H}}{H}=\frac{\hat{W}}{W}=\frac{1}{32}$. } consists of $v_{\hat{p}}$, $\tilde{F}\in\mathbb{R}^{\tilde{C}\times \tilde{H}\times \tilde{W}}$\footnote{$\tilde{C}$ denotes the number of channels of $v_{\tilde{p}}$. $\frac{\tilde{H}}{H}=\frac{\tilde{W}}{W}=\frac{1}{4}$.} consists of $v_{\tilde{p}}$, and $f_{\theta_2}$ is the lightweight Flattening Module which will be elaborated below. Model parameters $\theta_1$, $\theta_2$ and $\theta_3$ are updated via backpropagation \cite{rumelhart1988learning}.

\subsection{Flattening Module}
The proposed Flattening Module is illustrated in Fig.\ref{fig:fig2}, that takes as input $\hat{F}$ and then outputs $\tilde{F}$, allowing it to be seamlessly integrated into the FCN framework. More specifically, the coarse-grained feature maps $\hat{F}$ are fed into the depthwise separable group convolution (DWSGConv) layer, every time performing a convolution operation, outputting a grid of pixelwise visual descriptors (e.g., $8\times 8$), denoted by $G_{\bar{i}, \bar{j}}=\lbrace v_{\tilde{p}}:\: \lfloor\frac{\tilde{i}}{s_2}\rfloor=\bar{i}, \: \lfloor\frac{\tilde{j}}{s_2}\rfloor=\bar{j}\rbrace$ where $\bar{i}, \bar{j}\in\lbrace 0,\ldots,s_2-1\rbrace$, which are stacked initially along channel dimension. Then, the set of pixelwise visual descriptors corresponding to each single grid, i.e., $G_{\bar{i}, \bar{j}}$, are shifted from channel dimension to spatial domain and arranged from top left to bottom right. Next we go into details about two core components: the DWSGConv layer and the rearrangement operator.

\subsubsection{Depthwise Separable Group Convolution Layer}
Assume that a regular convolution whose kernel is of size $(4096=64\times 8\times 8)\times 3\times 3$ (followed by a batch normalization \cite{ioffe2015batch} and a ReLU activation \cite{nair2010rectified}) is opted to implement $f_{\theta_2}$, where $64$ denotes the number of channels $\tilde{C}$ of each $v_{\tilde{p}}$ in a $8\times 8$ grid. Since the number of channels of the feature maps produced by a FCN is typically large (e.g., $2048$ for Bottleneck-based ResNet), the single layer would cost a huge amount of parameters ($\sim75$M), which is clearly unreasonable. 

\begin{figure}[h!]
\centering\includegraphics[scale=1]{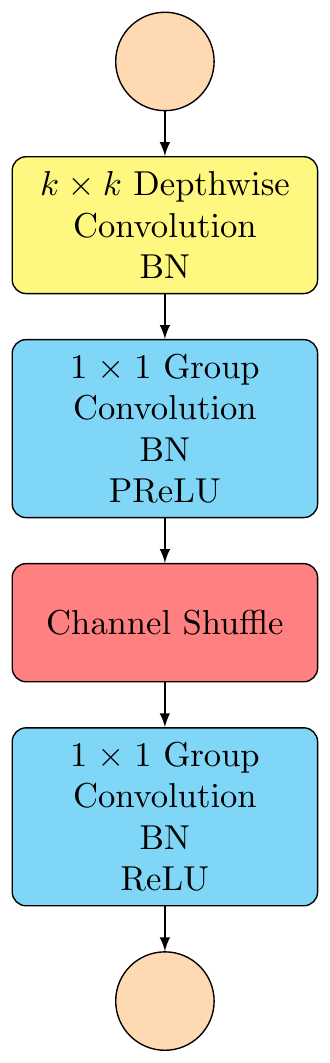}
\caption{Depthwise separable group convolution layer. Factorize a standard convolution layer with BN and ReLU into four separate components. BN: batch normalization. PReLU: Parametric ReLU.}
\label{fig:fig3}
\end{figure}

Inspired by \cite{zhang2018shufflenet}, \cite{sun2018igcv3}, \cite{xie2018interleaved}, \cite{zhang2017interleaved}, we propose a novel layer module based on the factorized convolution, termed depthwise separable group convolution (DWSGConv) layer, in order to output features for a grid of pixel locations simultaneously, without incurring a dramatic increase in the number of parameters. As illustrated in Fig. \ref{fig:fig3}, it can be divided into four components. The first component is a computationally economical $k\times k$ depthwise convolution \cite{chollet2017xception} followed by a batch normalization. The second component is composed of a pointwise group convolution, a batch normalization, and a Parametric ReLU (PReLU) activation \cite{he2015delving}. The third component is a channel shuffle operator to enable information communication between different groups of channels and improve accuracy \cite{zhang2018shufflenet}. The last component is another pointwise group convolution followed by a batch normalization and a ReLU activation. The ensemble of four components acts as an effective and efficient alternative of a regular convolution layer. Moreover, we have found empirically that inserting PReLU activation improves accuracy in some cases.

Many state-of-the-art neural network designs \cite{zhang2018shufflenet}, \cite{sun2018igcv3}, \cite{xie2018interleaved}, \cite{zhang2017interleaved}, \cite{howard2017mobilenets}, \cite{sandler2018mobilenetv2}, \cite{ma2018shufflenet}  incorporate depthwise separable convolution, pointwise group convolution, and channel shuffle operation into their building blocks to reduce the computational cost and the number of parameters while maintaining similar performance. Besides, the proposed DWSGConv layer is similar to the IGCV$3$ block proposed in \cite{sun2018igcv3}. In contrast to these works aiming at building lightweight and efficient models for mobile applications, the purpose of the proposed DWSGConv layer is to efficiently convert patchwise visual descriptors to pixelwise visual descriptors, preventing the difficulty in optimization.

\subsubsection{Rearranging Pixelwise Visual Descriptors}
\begin{figure}[h!]
\centering\includegraphics[width=0.49\textwidth]{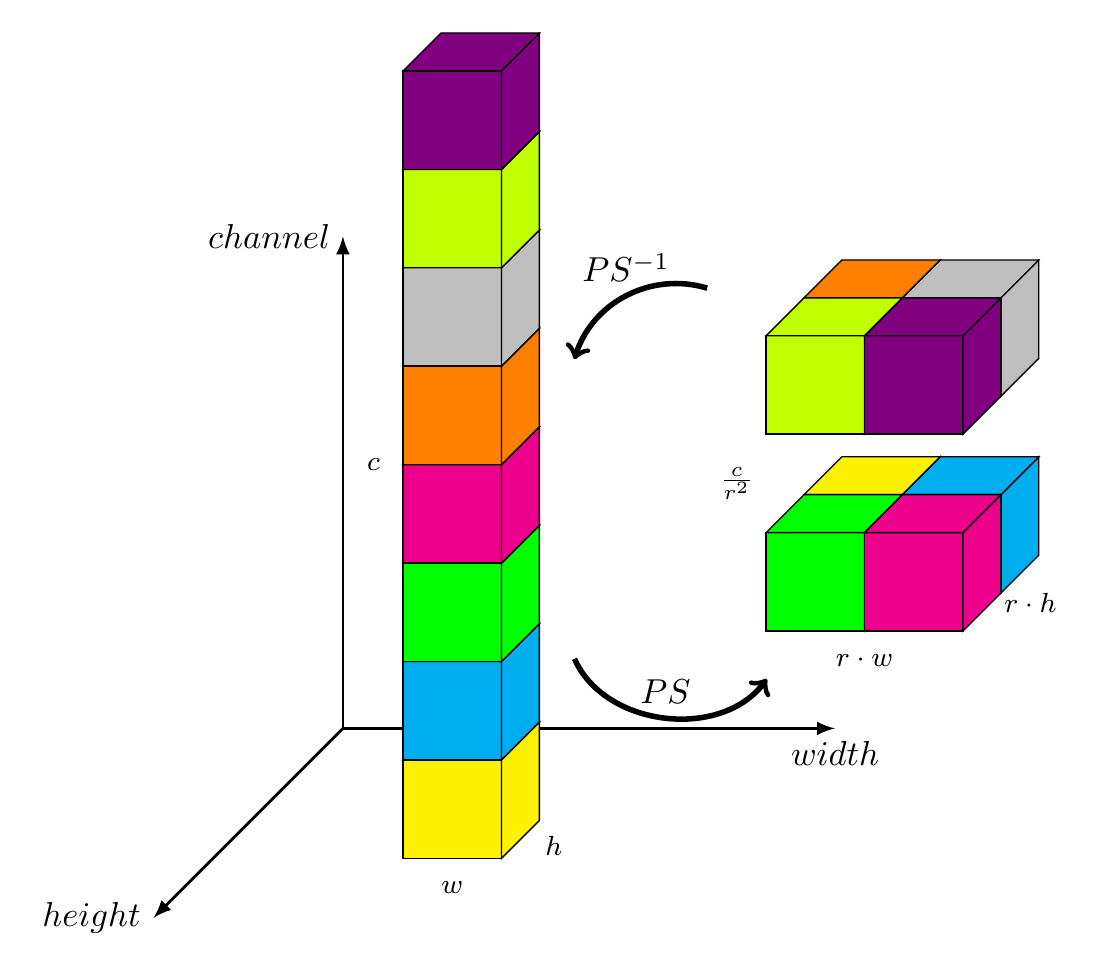}
\caption{An illustration of pixel shuffle operation. For presentation clarity, we only consider the case where the tensor is of shape $c=8, h=1, w=1$ and the upscaling factor $r$ is $2$. (a) is the original tensor before pixel shuffle. (b) is the tensor after pixel shuffle. We deliberately separate the tensor in (b) along channel dimension for better visualization. The same element before and after pixel shuffle is shown in the same color. $PS$: pixel shuffle operation.}
\label{fig:fig4}
\end{figure}

Pixel shuffle operator was first introduced in \cite{shi2016real}, termed periodic shuffling, as a component of the sub-pixel convolution layer, for the purpose of implementing transposed convolution \cite{dumoulin2016guide} efficiently. In our work, pixel shuffle operator is regarded as a bijection function that alters nothing but the $3$D coordinates of elements in a tensor. Fig. \ref{fig:fig4} illustrates a specific design, implemented in PyTorch \cite{paszke2017automatic}, of pixel shuffle operator. Furthermore, it is clear from this illustration that pixel shuffle operator rearranges elements from depth to space in a deterministic fashion, hence being differentiable and allowing the whole network to be trained end-to-end. 

Pixel shuffle operator in combination with channel shuffle operator \cite{zhang2018shufflenet} with the group number set to $s_2^2$, denoted by $R$, can easily realize shifting $v_{\tilde{p}}\in G_{\bar{i}, \bar{j}}$ from channel dimension to spatial domain. Although we have found empirically that simply using pixel shuffle operation would not hurt performance, we still stick to the current practice to achieve conceptual clarity. 

The manner in which pixelwise visual descriptors are rearranged is similar to that of \cite{tian2019decoders}, where a data-dependent upsampling method, termed DUpsampling, was proposed to address the limitations caused by data-independent bilinear upsampling. The main difference between our work and DUpsamling lies in that the upsampling filters in DUpsampling are pre-computed under some metric to minimize the reconstruction error between the ground truths and the compressed ones, however, our method is trained end-to-end with the only loss function specified by the target task without involving such reconstruction procedure. Besides, we have not observed difficulties in optimization during the training stage, hence without the need of designing specific loss functions.  

\subsection{Reformulating the Flattening Module as a Nonlinear Predictor}
In what follows, we offer a different perspective to help in understanding the simplicity of our method, as illustrated in Fig.\ref{fig:fig2} along the dashed line. First we fold the ground truth label map up via the inverse transformation $R^{-1}$ instead of arranging the outputs of the DWSGConv layer from depth to space. Then we let $v_{\tilde{p}}$ produced by the DWSGConv layer directly go through a fully-connected layer, yielding the final outputs $Y_{\tilde{p}}$. The overall pipeline is mathematically formulated as follows:
{
\begin{align}
Data\:Preparation&:\:\begin{cases}\dot{Y}_{\tilde{p}}=downsample\left(\dot{Y}_p\right)\\
T=R^{-1}\left(\dot{Y}_{\tilde{p}}\right)\\
\end{cases}\label{eq:eq5}\\
\hat{F}&=FCN\left(X\right)\label{eq:eq6}\\
g&=fc\circ DWSGConv\label{eq:eq7}\\
P&=g\left(\hat{F}\right)\label{eq:eq8}\\
\ell&=Loss\left(P,\:T\right)\label{eq:eq9}\\
\nonumber
\end{align}
}%
where $\dot{Y}_p\in\mathbb{R}^{C\times H\times W}$ denotes the ground truth label map, $C$ denotes the number of class labels or real valued outputs, $T\in\mathbb{R}^{C\cdot s_2^2\times \hat{H}\times \hat{W}}$ denotes the folded target, $fc$ denotes a fully-connected layer, and $P\in\mathbb{R}^{C\cdot s_2^2\times \hat{H}\times \hat{W}}$ denotes the final prediction. Note that $Y_{\tilde{p}}=R\left(P\right)$ under the same set of model parameters, i.e., this alternative procedure is completely equivalent to the original pipeline. Therefore, in order to obtain the training loss $\ell$, we only need to take a single step of feeding the coarse-grained feature maps $\hat{F}$ outputted by a FCN into the nonlinear predictor $g$, behaving just like a regular classification network. From this viewpoint, it is reasonable to consider that our proposed FlatteNet is a decoding-free approach. Although dilated convolution based methods similarly output predictions directly using the feature maps produced by a FCN (dilated-FCN), our method is more efficient in terms of the computational complexity owing to the usage of coarse-grained feature maps generated via a FCN without removing any subsampling operations.

\section{Experiments and Analysis}
\label{sec:experimentsandanalysis}
To evaluate the proposed Flattening Module, we carry out comprehensive experiments on MPII Human Pose dataset \cite{andriluka2014}, PASCAL-Context dataset \cite{mottaghi2014}, and PASCAL VOC dataset \cite{everingham2015}. Empirical results demonstrate that the Flattening Module based network (FlatteNet) achieves comparable or slightly better performance compared to its dilated convolution based or decoder based counterpart across three benchmark tasks with much fewer parameters and lower computational complexity. Our experiments are implemented with PyTorch \cite{paszke2017automatic}.

\subsection{HUMAN POSE ESTIMATION}
\subsubsection{Dataset}
MPII \cite{andriluka2014} is the benchmark dataset for single person 2D pose estimation. The images were collected from YouTube videos, covering daily human activities with complex poses and image appearances. There are about $25K$ images. In total, about $29K$ annotated poses are for training and another $7K$ are for testing.

\subsubsection{Training}
We use the state-of-the-art ResNet \cite{he2016deep} as backbone network. It is pre-trained on ImageNet classification dataset \cite{deng2009imagenet}. Adam \cite{kingma2014adam} is used for optimization. In the training for pose estimation, the base learning rate is $1\mathrm{e}{-3}$. It drops to $1\mathrm{e}{-4}$ at $90$ epochs and $1\mathrm{e}{-5}$ at $120$ epochs. There are $140$ epochs in total. Mini-batch size is $32$. A single GTX1080Ti GPU is used. Data augmentation includes random rotation ($\pm 30$ degrees), scaling ($\pm 25\%$) and flip. The input image is normalized to $256\times 256$. The set of hyperparameters related to the Flattening Module is set to the values shown in Table \ref{tab:table1}, unless otherwise specified. For PReLU activation, we choose the channel-wise version and set initial values to $1$. Note the values of $g_1$, $g_2$ and $g_3$ are set according to the complementary condition proposed in \cite{xie2018interleaved}.

\begin{table}[h!]
\caption{The default hyperparameter setting of the Flattening Module. DWConv: depthwise convolution. FPGConv: the first pointwise group convolution. SPGConv: the second pointwise group convolution. CS: channel shuffle operation. PS: pixel shuffle operation. $k$: kernel size. $g$: group number. $u$: upscaling factor. $c_{in}$: the number of input channels. $c_{out}$: the number of output channels. $s$: stride.}
\label{tab:table1}
\renewcommand{\arraystretch}{1.3}
\setlength{\tabcolsep}{3.0pt}
\footnotesize
\centering
\begin{tabular}{@{}cccccc@{}} \toprule
\multirow{2}{*}{DWConv} & \multirow{2}{*}{FPGConv} & \multirow{2}{*}{CS} & \multirow{2}{*}{SPGConv} & \multicolumn{2}{c}{Rearrangement}\\ \cmidrule{5-6}
&&&& CS & PS\\ \midrule
$k=3$ & $g_1=32$ & \multirow{2}{*}{$g_2=32$} & $g_3=64$ & \multirow{2}{*}{$g_4=8^2$} & \multirow{2}{*}{$u=8$}\\
$s=1$& $c_{out}=c_{in}$ & & $c_{out}=1\times c_{in}$&\\
\bottomrule
\end{tabular}
\end{table}

\subsubsection{Evaluation}
For performance evaluation, MPII \cite{andriluka2014} uses PCKh metric, which is the percentage of correct keypoint. A keypoint is correct if its distance to the ground truth is less than a fraction $\alpha$ of the head segment length. The metric is denoted as PCKh@$\alpha$.

Commonly, PCKh@$0.5$ metric is used for comparison \cite{mpii}. For evaluation under high localization accuracy, we also report PCKh@$0.1$ and AUC (area under curve, the averaged PCKh when $\alpha$ varies from $0$ to $0.5$).

Since the annotation on test set is not available, all our ablation studies are evaluated on an about $3k$ validation set which is separated out from the training set, following previous common practice \cite{newell2016stacked}. Training is performed on the remaining training data.

Following the common practice in \cite{chen2018cascaded} \cite{newell2016stacked}, the joint prediction is predicted on the averaged heatmaps of the original and flipped images. A quarter offset in the direction from highest response to the second highest response is used to obtain the final location.

\begin{table}[h!]
\caption{Ablation study of the DWSGConv layer on MPII validation dataset using ResNet-$50$ as the backbone network. RandomPerm: a random permutation along channel dimension. Note that \#Params only counts the DWSGConv layer.}
\label{tab:table2}
\renewcommand{\arraystretch}{1.3}
\setlength{\tabcolsep}{3.0pt}
\scriptsize
\centering
\begin{tabular}{@{}lSSSS@{}} \toprule
Design Options & {PCKh@0.5} & {PCKh@0.1} & AUC & {\#Params(M)}\\ \midrule
Regular $1\times 1$ conv & 88.59 & 33.34 & 59.29 & 4.19\\
PReLU & 88.60 & 33.11 & 59.18 & 0.23\\
ReLU & 85.73 & 31.44 & 56.93 & 0.23\\
SPGConv $c_{out}=8\times c_{in}$ & 88.39 & 33.00 & 59.10 & 0.71\\
Rearrangement$=$PS & 88.41 & 33.07 & 59.02 & 0.23\\
Rearrangement$=$RandPerm + PS & 88.29 & 32.70 & 58.89 & 0.23\\
\bottomrule
\end{tabular}
\end{table}

\begin{figure*}[t!]
\centering\includegraphics[width=\textwidth]{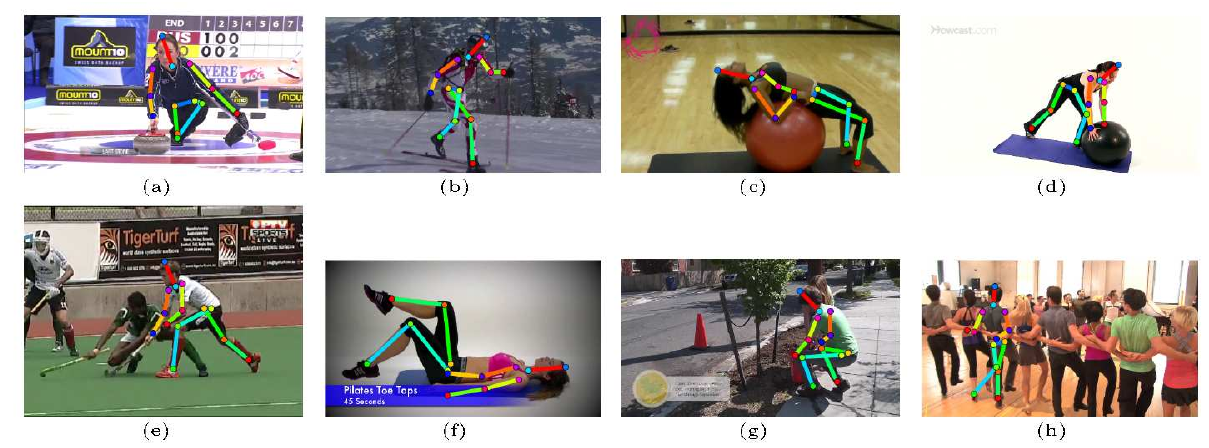}
\caption{Qualitative results of FlatteNet on MPII test dataset. It can be seen that both our method produces good predictions in most cases.}
\label{fig:fig5}
\end{figure*}

\subsubsection{Ablation study}
We firstly explore several design choices of the Flattening Module\footnote{Note that the rearrangement component contains no parameter.} and the results are shown in Table \ref{tab:table2}. Compared to the regular $1\times 1$ convolution, the DWSGConv layer yields nearly identical results under all three metrics with much fewer parameters. When the number of output channels is increased by a factor of $8$, the accuracy barely drops, indicating that the Flattening Module can efficiently handle the enormous output space inherent in the dense pixelwise prediction problem. Besides, it can be observed that replacing PReLU activation with ReLU activation leads to a significant drop in accuracy ($2.87 \downarrow$). Finally we compare three different methods of rearranging elements from channel dimension to spatial domain: (i) channel shuffle followed by pixel shuffle (the default setting); (ii) only pixel shuffle; (iii) random permutation followed by pixel shuffle. From the results, shown in Table \ref{tab:table2}, we find that there is no significant difference in accuracy, since the DWSGConv layer is dense due to strictly following the complementary condition proposed in \cite{xie2018interleaved}. For conceptual clarity, we stick to the default setting.

\begin{table}[h!]
\caption{Comparison of model complexity on MPII validation dataset.}
\label{tab:table3}
\renewcommand{\arraystretch}{1.3}
\setlength{\tabcolsep}{3.0pt}
\footnotesize
\centering
\begin{tabular}{@{}lSSSSS@{}} \toprule
Backbone & {PCKh@0.5} & {PCKh@0.1} & AUC & {\#Params(M)} & GFLOPs\\ \midrule
ShuffleNet~v2 & 84.25 & 24.81 & 52.50 & 1.42 &0.19\\
ResNet-$50$ & 88.60 & 33.11 & 59.18 & 23.77 & 4.99\\
ResNet-$101$ & 88.91 & 33.42 & 59.51 & 42.73 & 9.50\\
\bottomrule
\end{tabular}
\end{table}

Table \ref{tab:table3} shows the results of the Flattening Module combined with different backbones. It is clear that using a network with large capacity improves the accuracy. Besides, the performance and the complexity of the whole system largely depends on the backbone network, partly supporting our viewpoint of regarding the Flattening Module as a nonlinear predictor. It is worth noting that using ShuffleNet-v2 $1\times$ \cite{ma2018shufflenet} as the backbone architecture \footnote{Note that $c_{out}=2\times c_{in}$ in the SPGConv.} achieves respectable performance of approximately $84\%$ on PCKh@0.5. Developing highly computation-efficient CNN architectures to carry out dense prediction tasks especially for mobile devices has been hampered by the computationally expensive operations, such as dilated convolution and transposed convolution, which are indispensable to the current dominant dense prediction frameworks. On the contrary, our method can directly benefit from efficient encoder designs which have been extensively studied\cite{zhang2018shufflenet}, \cite{howard2017mobilenets}, \cite{sandler2018mobilenetv2}, \cite{ma2018shufflenet}, \cite{huang2018condensenet}. We hope that our method would help inspiring new ideas of efficient network architecture designs for dense prediction tasks on mobile platforms.

\begin{table}[h!]
\caption{Overall Comparison of several state-of-the-art methods and ours on MPII validation dataset.}
\label{tab:table4}
\renewcommand{\arraystretch}{1.3}
\setlength{\tabcolsep}{3.0pt}
\footnotesize
\centering
\begin{tabular}{@{}lcSSS@{}} \toprule
Method & Backbone & {PCKh@0.5}  & {\#Params(M)} & GFLOPs\\ \midrule
FlatteNet & ResNet-$50$ & 88.6 & 24 & 5.0\\
Simple~Baseline \cite{xiao2018simple} & ResNet-$50$ & 88.5 & 34 & 12.0\\
Integral~Reg \cite{sun2018integral} & ResNet-$50$& 87.3 & 26 & 6.2\\
Hourglass-$8$ \cite{newell2016stacked} & {-} & 88.1 & 51 & 25.6\\
\bottomrule
\end{tabular}
\end{table}

Table \ref{tab:table4} further compares the accuracy, model size, and computational complexity trade-off between our approach and several representative methods. Our approach performs on par with theses methods, but the number of parameters and FLOPs are significantly lower. In particular, compared to Simple~Baseline \cite{xiao2018simple} which has achieved state-of-the-art performance on challenging benchmark datasets \cite{andriluka2014}, \cite{lin2014microsoft}, \cite{andriluka2018posetrack}, our approach further improves efficiency ($29\%\downarrow$ on \#Params and $58\%\downarrow$ on GFLOPs) while keeping the same accuracy. Compared to the integral regression method \cite{sun2018integral}, which is proposed to reduce the computational cost incurred by producing high-resolution heatmaps, heatmap based FlatteNet still uses relatively fewer parameters and smaller computational cost. Finally it is reasonable to expect that using multi-stage architecture can further boost the performance of our method.

\begin{table}[h!]
\caption{Comparison of the number of subsampling operations on MPII validation dataset. The specific hyperparameter settings are shown in Table \ref{tab:table11}.}
\label{tab:table5}
\renewcommand{\arraystretch}{1.3}
\setlength{\tabcolsep}{3.0pt}
\fontsize{7.2}{7.8}\selectfont
\centering
\begin{tabular}{@{}ccSSSS@{}} \toprule
{\#Subsampling} & Resolution & {PCKh@0.5} & {PCKh@0.1} & AUC & {\#Params(M)}\\ \midrule
5 & $8\times 8$ & 88.60 & 33.11 & 59.18 & 23.77\\
6 & $4\times 4$ & 88.50 & 32.72 & 58.87 & 23.95\\
7 & $2\times 2$ & 88.17 & 31.77 & 58.27 & 27.27\\
\bottomrule
\end{tabular}
\end{table}

The lightweight property of the proposed Flattening Module allows us to investigate the relationship between the prediction accuracy and the number of subsampling operations without concern for difficulty in optimization. From the results, shown in Table \ref{tab:table5}, there is no significant drop in accuracy with the growth of the number of subsampling operations, which appears to be in conflict with the conventional opinion that the subsampling operations are detrimental to dense prediction. We hope our preliminary experiments would encourage the community to reconsider the role of subsampling operations in dense prediction.

\subsubsection{Results on the MPII test set}
Table \ref{tab:table6} presents the PCKh@$0.5$ results of our method as well as state-of-the-art methods on the MPII test set. Note that our FlatteNet achieves $91.3$ PCKh@$0.5$ score using ResNet-$152$ as backbone and the input resolution is set to $384\times 384$. Our intent is not to push the state-of-the-art results, but to demonstrate the effectiveness of our approach.

\begin{table}[h!]
\caption{Comparison to state-of-the-art methods on the MPII test set.}
\label{tab:table6}
\renewcommand{\arraystretch}{1.3}
\setlength{\tabcolsep}{3.0pt}
\fontsize{6.2}{7.2}\selectfont
\centering
\begin{tabular}{@{}lSSSSSSSS@{}} \toprule
{Method} & {Head} & {Shoulder} & {Elbow} & {Wrist} & {Hip} & {Knee} & {Ankle} & {Total}\\ \midrule
Insafutdinov \cite{insafutdinov2016deepercut} & 96.8  & 95.2  & 89.3  & 84.4  & 88.4  & 83.4 & 78.0 & 88.5\\
Sun \cite{sun2018integral} & 98.1 & 96.0 & 90.4 & 85.7 & 90.1 & 85.8 & 81.0 & 90.0\\
Newell \cite{newell2016stacked} & 98.2  & 96.3  & 91.2  & 87.1  & 90.1  & 87.4 & 83.6 & 90.9\\
Ning \cite{ning2017knowledge} & 98.1  & 96.3  & 92.2  & 87.8  & 90.6  & 87.6 & 82.7 & 91.2 \\
Luvizon \cite{luvizon2019human} & 98.1  & 96.6  & 92.0  & 87.5  & 90.6  & 88.0 & 82.7 & 91.2\\
Chu \cite{chu2017multi} & 98.5  & 96.3  & 91.9  & 88.1  & 90.6  & 88.0 & 85.0 & 91.5 \\
Chou \cite{chou2018self} & 98.2  & 96.8  & 92.2  & 88.0  & 91.3  & 89.1 & 84.9 & 91.8 \\
Chen \cite{chen2017adversarial} & 98.1  & 96.5  & 92.5  & 88.5  & 90.2  & 89.6 & 86.0 & 91.9 \\
Yang \cite{yang2017learning} & 98.5  & 96.7  & 92.5  & 88.7  & 91.1  & 88.6 & 86.0 & 92.0 \\
Tang \cite{tang2018deeply} & 98.4  & 96.9  & 92.6  & 88.7  & 91.8  & 89.4 & 86.2 & 92.3\\
\hline
Simple~Baseline \cite{xiao2018simple} & 98.5 & 96.6 & 91.9 & 87.6 & 91.1 & 88.1 & 84.1 & 91.5\\
FlatteNet & 98.4 & 96.4 & 91.8 & 87.7 & 90.6 & 87.9 & 83.3 & 91.3\\
\bottomrule
\end{tabular}
\end{table}

\subsubsection{Qualitative results}
Qualitative results of FlatteNet are shown in Fig. \ref{fig:fig5}. From the qualitative results of some example images, it can be clearly seen that our approach is robust to large variability of body appearances, severe body deformation, and changes in viewpoint. However, it suffers from the common shortcomings such as having difficulty in dealing with occlusion or self-occlusion, e.g., the predictions in subfigure (h).

\begin{figure*}[t!]
\centering\includegraphics[width=\textwidth]{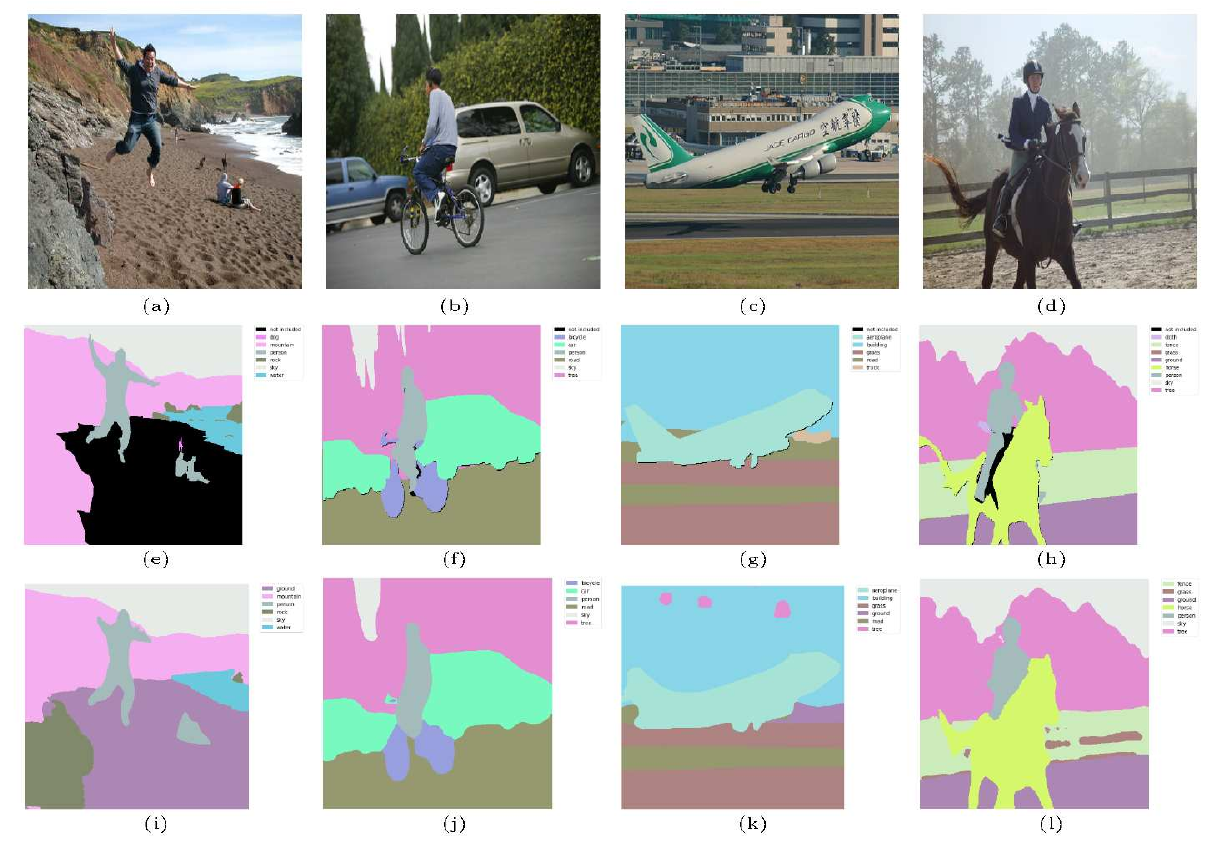}
\caption{Qualitative results of FlatteNet on PASCAL-Context validation dataset. The input images and ground truths are shown in the first row and the second row, respectively. The corresponding segmentation results outputted by FlatteNet are shown in the last row.}
\label{fig:fig6}
\end{figure*}

\subsection{Semantic Segmentation}
\subsubsection{Dataset}
The PASCAL context dataset \cite{mottaghi2014} includes $4988$ scene images for training and $5105$ images for testing with $59$ semantic labels and $1$ background label.

\subsubsection{Training}
The data are augmented by random cropping, random scaling in the range of $\left[0.5, 2\right]$, and random horizontal flipping. Following the widely-used training strategy \cite{zhang2018context}, we resize the images to $480\times 480$ and use the SGD optimizer with base learning rate $4\mathrm{e}{-3}$, the momentum of $0.9$, and the weight decay of $1\mathrm{e}{-4}$. The poly learning rate policy with the power of $0.9$ is used for dropping the learning rate. The set of hyperparameters related to the Flattening Module is set to the values shown in Table \ref{tab:table7}. Note that we use two DWSGConv layers to better exploit the context. All the models are trained for $100$ epochs with batchsize of $8$ on single GTX1080Ti GPU, hence using BN \cite{ioffe2015batch} instead of Synchronize BN \cite{zhang2018context}.

\begin{table}[h!]
\caption{The default hyperparameter setting of the Flattening Module used in semantic segmentation task.}
\label{tab:table7}
\renewcommand{\arraystretch}{1.3}
\setlength{\tabcolsep}{3.0pt}
\fontsize{5.8}{7.3}\selectfont
\centering
\begin{tabular}{@{}ccccccc@{}} \toprule
\multirow{2}{*}{DWConv} & \multirow{2}{*}{FPGConv} & \multirow{2}{*}{CS} & \multirow{2}{*}{SPGConv} & \multicolumn{2}{c}{Rearrangement} & \multirow{2}{*}{\#Params(M)}\\ \cmidrule{5-6}
&&&& CS & PS\\ \midrule
$k=7$ & $g_1=32$ & \multirow{2}{*}{$g_2=32$} & $g_3=64$ & \multirow{4}{*}{$g_4=8^2$} & \multirow{4}{*}{$u=8$} & \multirow{4}{*}{1.40}\\
$s=1$& $c_{out}=c_{in}$ & & $c_{out}=2\times c_{in}$&\\
$k=7$ & $g_1=64$ & \multirow{2}{*}{$g_2=64$} & $g_3=64$ &\\
$s=1$ & $c_{out}=c_{in}$ & & $c_{out}=2\times c_{in}$&\\
\bottomrule
\end{tabular}
\end{table}

\subsubsection{Evaluation}
Following standard testing procedure \cite{zhang2018context}, the models are tested with the input size of $480\times 480$. We use standard evaluation metric of mean Intersection of Union (mIoU) for $59$ classes without background and evaluate our approach and other methods using six scales of $\left[0.5, 0.75, 1, 1.25, 1.5\right]$ and flipping.

\begin{table}[h!]
\caption{Semantic segmentation results on PASCAL-Context. The methods are evaluated on $59$ classes. S: single scale evaluation.}
\label{tab:table8}
\renewcommand{\arraystretch}{1.3}
\setlength{\tabcolsep}{3.0pt}
\footnotesize
\centering
\begin{tabular}{@{}lcSSS@{}} \toprule
Method & Backbone & mIoU & {\#Params(M)} & GFLOPs\\ \midrule
UNet++ \cite{zhou2018unet++} & {-} & 47.7 & 36.6 & 452.1\\
PSPNet \cite{zhao2017pyramid} & Dilated-ResNet-$101$ & 47.8 & 68.1 & 222.3\\
EncNet \cite{zhang2018context} & Dilated-ResNet-$101$ & 52.6 & 62.8 & 212.0\\
\hline
FlatteNet-S & \multirow{2}{*}{ResNet-$101$} & 47.3 & {\multirow{2}{*}{43.5}} & {\multirow{2}{*}{33.6}}\\
FlatteNet &  & 48.8 & & \\
\bottomrule
\end{tabular}
\end{table}

\begin{figure*}[t!]
\centering\includegraphics[width=\textwidth]{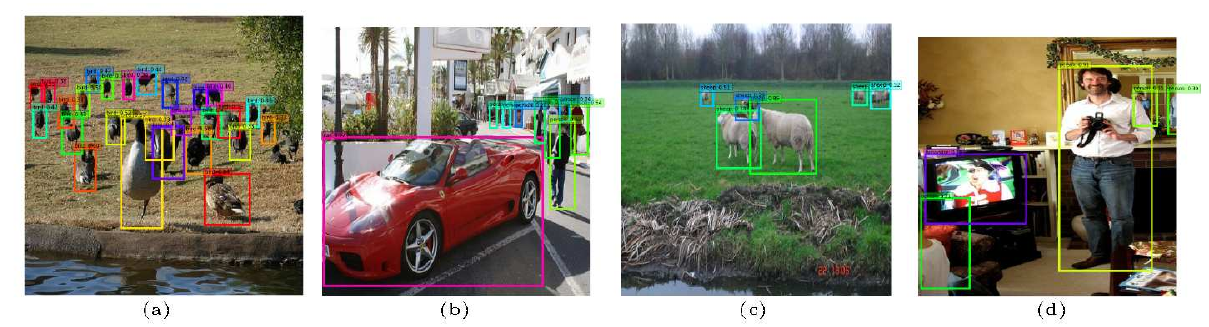}
\caption{Qualitative results of FlatteNet on PASCAL VOC 2007 test set. Note that the example images are picked specially for highlighting the characteristics of keypoint based object detectors.}
\label{fig:fig7}
\end{figure*}

\subsubsection{Results}
To further demonstrate the effectiveness and generality of our method, we train FlatteNet for semantic segmentation on the PASCAL-Context dataset. Fig. \ref{fig:fig6} shows a few visual examples on validation set of PASCAL-Context. Quantitative results are shown in Table \ref{tab:table8}. Note that we do not adopt intermediate supervision and online hard example mining for training, which are used in three other methods. PSPNet employs four spatial pyramid pooling layers in parallel to exploit the global context information. UNet++ is an enhanced version of UNet \cite{ronneberger2015u} where the encoder and decoder sub-networks are connected through a series of nested, dense skip paths. EncNet employs a specially designed Context Encoding Module and Semantic Encoding Loss to capture global context and selectively highlight the class-dependent feature maps. However, our method, simply appending a factorized convolutional layer (Flattening Module) after the backbone network with the number of parameters $1.4$M, rivals the results of PSPNet and UNet++, clearly demonstrating its powerfulness. Despite performing worse than EncNet, we would like to point out that our method is orthogonal to the Context Encoding Module and other context aggregation modules, such as Object Context Pooling \cite{yuan2018ocnet}, Position Attention Module \cite{chen2017dual}, and Channel Attention Module \cite{chen2017dual}, and can further benefit from combining these modules. More importantly, our method achieves a substantial reduction in computational complexity (at least $6\times$). Besides, our method requires much fewer parameters compared to PSPNet and EncNet ($30\%\downarrow$). We argue that the impressive performance of our simple FlatteNet can be attributed to not removing any subsampling operations, which is consistent with our previous finding.

\subsection{OBJECT DETECTION}
\subsubsection{Dataset} 
PASCAL VOC \cite{everingham2015} is a popular object detection dataset. We train on VOC 2007 and VOC 2012 trainval sets, and test on VOC 2007 test set. It contains $16551$ training images and $4962$ testing images of $20$ categories.

\subsubsection{Object detection by keypoint estimation} 
Benefiting from the recent advance in the research area, which is reformulating object detection in a pixelwise prediction fashion, FlatteNet can be immediately extended to solve object detection task with only minor modifications on \textit{head} network. From among several models belonging to the family of keypoint based object detectors \cite{law2018cornernet}, \cite{tian2019fcos}, \cite{zhou2019objects}, \cite{duan2019centernet}, we adopt the method presented in \cite{zhou2019objects} to adapt FlatteNet due to its simplicity and competitive performance. 

\subsubsection{Training and testing}
During training and testing, we fix the input resolution to $512\times 512$. We use ResNet-$101$ as the backbone architecture. We use Adam \cite{kingma2014adam} to optimize the overall objective with initial learning rate $1.25\mathrm{e}{-4}$ dropped $10\times$ at $45$ and $60$ epochs, respectively. We use random flip, random scaling in the range of $\left[0.6, 1.4\right]$, random cropping and color jittering as data augmentation. We employ $3\times 3$ deformable convolution \cite{zhu2019deformable} in the last stage of ResNet-$101$ for fair comparison. The set of hyperparameters related to the Flattening Module is set to the values shown in Table \ref{tab:table9}. All the models are trained for $70$ epochs with batchsize of $8$ on single GTX1080Ti GPU. The evaluation metric is mean average precision (mAP) at IOU threshold $0.5$.

\begin{table}[h!]
\caption{The default hyperparameter setting of the Flattening Module used for object detection.}
\label{tab:table9}
\renewcommand{\arraystretch}{1.3}
\setlength{\tabcolsep}{3.0pt}
\footnotesize
\centering
\begin{tabular}{@{}cccccc@{}} \toprule
\multirow{2}{*}{DWConv} & \multirow{2}{*}{FPGConv} & \multirow{2}{*}{CS} & \multirow{2}{*}{SPGConv} & \multicolumn{2}{c}{Rearrangement}\\ \cmidrule{5-6}
&&&& CS & PS\\ \midrule
$k=5$ & $g_1=32$ & \multirow{2}{*}{$g_2=32$} & $g_3=32$ & \multirow{2}{*}{$g_4=8^2$} & \multirow{2}{*}{$u=8$}\\
$s=1$& $c_{out}=c_{in}$ & & $c_{out}=2\times c_{in}$&\\
\bottomrule
\end{tabular}
\end{table}

\subsubsection{Results}
Qualitative results are shown in Fig. \ref{fig:fig7}. Note that the example images are picked specially for highlighting the characteristics of keypoint based object detectors. As shown in the most of example images, they are good at dealing with small objects and crowded scenes. The potential problem of keypoint based object detectors is center point collision, which means two different objects might share the same center, if they perfectly align. In this scenario, keypoint based object detectors would only detect one of them, as shown in the subfigure (d), the conspicuous person inside of a tv-monitor is not detected. It is evident from the above discussion that FlatteNet behaves exactly the same as a regular keypoint based detector just much simpler.

Quantitative results are shown in Table \ref{tab:table10}. To save computation, CenterNet \cite{zhou2019objects} modifies the encoder-decoder architecture used in Simple~Baseline \cite{xiao2018simple} by changing the channels of the three transposed convolutional layers to $256$, $128$, $64$, respectively. Despite the modifications mentioned above, our method still requires fewer parameters and smaller computational cost while performing comparably with CenterNet. We also would like to mention that all the hyper-parameters are set according to CenterNet and not specifically finetuned for our network. Furthermore, it is reasonable to expect that the proposed Flattening Module can also be used in other keypoint based object detectors to reduce the number of parameters and computational cost while keeping performance intact. 

\begin{table}[h!]
\caption{Experimental results on Pascal VOC 2007 test set. Flip test is used for all models.}
\label{tab:table10}
\renewcommand{\arraystretch}{1.3}
\setlength{\tabcolsep}{3.0pt}
\small
\centering
\begin{tabular}{@{}llSSS@{}} \toprule
method & backbone & {mAP@0.5} & {\#Params(M)} & GFLOPs\\ \midrule
CenterNet \cite{zhou2019objects} & ResNet-$101$ & 78.7 & 44.6 & 43.0\\
FlatteNet & ResNet-$101$ & 78.2 & 43.8 & 39.9\\
\bottomrule
\end{tabular}
\end{table}

\section{Conclusion and Future Works}
\label{conclusionandfutureworks}
In this paper, we have proposed a novel scheme to produce high-resolution predictions by employing the simple and lightweight Flattening Module, in an effort to streamline the current dense prediction procedures. As shown in experiments, a common backbone network combined with the Flattening Module achieves comparable accuracy compared to its dilated convolution based or decoder based counterpart while only requiring a fraction of model size and computational cost. Given its effectiveness and efficiency, we hope FlatteNet can serve as a simple and strong alternative of current mainstream dense prediction networks. We also hope that our work can facilitate the study on efficient model designs for dense prediction tasks on embedded devices.

It remains unclear how to leverage the recent advances in the field of dense prediction within our simplified framework to attain state-of-the-art performance. Besides, the applications to other benchmark datasets \cite{lin2014microsoft}, \cite{andriluka2018posetrack}, \cite{cordts2016cityscapes} and dense prediction tasks, e.g., face landmark detection and human parsing, are important works in the future.

\appendices
\section{}
\begin{table}[h]
\caption{The hyperparameter settings of the Flattening Module used in Table \ref{tab:table5}.}
\label{tab:table11}
\renewcommand{\arraystretch}{1.3}
\setlength{\tabcolsep}{3.0pt}
\fontsize{5.5}{7.5}\selectfont
\centering
\begin{tabular}{@{}ccccccc@{}} \toprule
\multirow{2}{*}{\#Subsampling} & \multirow{2}{*}{DWConv} & \multirow{2}{*}{FPGConv} & \multirow{2}{*}{CS} & \multirow{2}{*}{SPGConv} & \multicolumn{2}{c}{Rearrangement}\\ \cmidrule{6-7}
&&&&& CS & PS\\ \midrule
\multirow{2}{*}{$6$} & $k=3$ & $g_1=32$ & \multirow{2}{*}{$g_2=32$} & $g_3=64$ & \multirow{2}{*}{$g_4=16^2$} & \multirow{2}{*}{$u=16$}\\
 &$s=2$ & $c_{out}=c_{in}$ & & $c_{out}=4\times c_{in}$&\\
\hline
\multirow{4}{*}{$7$} & $k=3$ & $g_1=32$ & \multirow{2}{*}{$g_2=32$} & $g_3=64$ & \multirow{4}{*}{$g_4=32^2$} & \multirow{4}{*}{$u=32$}\\
 &$s=2$ & $c_{out}=c_{in}$ & & $c_{out}=4\times c_{in}$&\\
&$k=3$ & $g_1=64$ & \multirow{2}{*}{$g_2=64$} & $g_3=128$ & \\
 &$s=2$ & $c_{out}=c_{in}$ & & $c_{out}=4\times c_{in}$&\\
\bottomrule
\end{tabular}
\end{table}

{
\bibliographystyle{IEEEtran}
\bibliography{references}

\begin{thebibliography}{10}
\providecommand{\url}[1]{#1}
\csname url@samestyle\endcsname
\providecommand{\newblock}{\relax}
\providecommand{\bibinfo}[2]{#2}
\providecommand{\BIBentrySTDinterwordspacing}{\spaceskip=0pt\relax}
\providecommand{\BIBentryALTinterwordstretchfactor}{4}
\providecommand{\BIBentryALTinterwordspacing}{\spaceskip=\fontdimen2\font plus
\BIBentryALTinterwordstretchfactor\fontdimen3\font minus
  \fontdimen4\font\relax}
\providecommand{\BIBforeignlanguage}[2]{{%
\expandafter\ifx\csname l@#1\endcsname\relax
\typeout{** WARNING: IEEEtran.bst: No hyphenation pattern has been}%
\typeout{** loaded for the language `#1'. Using the pattern for}%
\typeout{** the default language instead.}%
\else
\language=\csname l@#1\endcsname
\fi
#2}}
\providecommand{\BIBdecl}{\relax}
\BIBdecl

\bibitem{chen2018encoder}
L.-C. Chen, Y.~Zhu, G.~Papandreou, F.~Schroff, and H.~Adam, ``Encoder-decoder
  with atrous separable convolution for semantic image segmentation,'' in
  \emph{Proceedings of the European conference on computer vision (ECCV)},
  2018, pp. 801--818.

\bibitem{xiao2018simple}
B.~Xiao, H.~Wu, and Y.~Wei, ``Simple baselines for human pose estimation and
  tracking,'' in \emph{Proceedings of the European Conference on Computer
  Vision (ECCV)}, 2018, pp. 466--481.

\bibitem{zhang2018progressive}
X.~Zhang, T.~Wang, J.~Qi, H.~Lu, and G.~Wang, ``Progressive attention guided
  recurrent network for salient object detection,'' in \emph{Proceedings of the
  IEEE Conference on Computer Vision and Pattern Recognition (CVPR)}, 2018, pp.
  714--722.

\bibitem{silberman2012indoor}
N.~Silberman, D.~Hoiem, P.~Kohli, and R.~Fergus, ``Indoor segmentation and
  support inference from rgbd images,'' in \emph{Proceedings of the European
  Conference on Computer Vision (ECCV)}.\hskip 1em plus 0.5em minus 0.4em\relax
  Springer, 2012, pp. 746--760.

\bibitem{ilg2017flownet}
E.~Ilg, N.~Mayer, T.~Saikia, M.~Keuper, A.~Dosovitskiy, and T.~Brox, ``Flownet
  2.0: Evolution of optical flow estimation with deep networks,'' in
  \emph{Proceedings of the IEEE conference on computer vision and pattern
  recognition (CVPR)}, 2017, pp. 2462--2470.

\bibitem{shi2016real}
W.~Shi, J.~Caballero, F.~Husz{\'a}r, J.~Totz, A.~P. Aitken, R.~Bishop,
  D.~Rueckert, and Z.~Wang, ``Real-time single image and video super-resolution
  using an efficient sub-pixel convolutional neural network,'' in
  \emph{Proceedings of the IEEE conference on computer vision and pattern
  recognition (CVPR)}, 2016, pp. 1874--1883.

\bibitem{isola2017image}
P.~Isola, J.-Y. Zhu, T.~Zhou, and A.~A. Efros, ``Image-to-image translation
  with conditional adversarial networks,'' in \emph{Proceedings of the IEEE
  Conference on Computer Vision and Pattern Recognition (CVPR)}, 2017, pp.
  1125--1134.

\bibitem{law2018cornernet}
H.~Law and J.~Deng, ``Cornernet: Detecting objects as paired keypoints,'' in
  \emph{Proceedings of the European Conference on Computer Vision (ECCV)},
  2018, pp. 734--750.

\bibitem{tian2019fcos}
Z.~Tian, C.~Shen, H.~Chen, and T.~He, ``Fcos: Fully convolutional one-stage
  object detection,'' \emph{arXiv preprint arXiv:1904.01355}, 2019.

\bibitem{zhou2019objects}
X.~Zhou, D.~Wang, and P.~Kr{\"a}henb{\"u}hl, ``Objects as points,'' \emph{arXiv
  preprint arXiv:1904.07850}, 2019.

\bibitem{duan2019centernet}
K.~Duan, S.~Bai, L.~Xie, H.~Qi, Q.~Huang, and Q.~Tian, ``Centernet: Keypoint
  triplets for object detection,'' \emph{arXiv preprint arXiv:1904.08189},
  2019.

\bibitem{long2015fully}
J.~Long, E.~Shelhamer, and T.~Darrell, ``Fully convolutional networks for
  semantic segmentation,'' in \emph{Proceedings of the IEEE conference on
  computer vision and pattern recognition (CVPR)}, 2015, pp. 3431--3440.

\bibitem{krizhevsky2012imagenet}
A.~Krizhevsky, I.~Sutskever, and G.~E. Hinton, ``Imagenet classification with
  deep convolutional neural networks,'' in \emph{Advances in neural information
  processing systems}, 2012, pp. 1097--1105.

\bibitem{simonyan2014very}
K.~Simonyan and A.~Zisserman, ``Very deep convolutional networks for
  large-scale image recognition,'' \emph{arXiv preprint arXiv:1409.1556}, 2014.

\bibitem{he2016deep}
K.~He, X.~Zhang, S.~Ren, and J.~Sun, ``Deep residual learning for image
  recognition,'' in \emph{Proceedings of the IEEE Conference on Computer Vision
  and Pattern Recognition (CVPR)}, 2016, pp. 770--778.

\bibitem{deng2009imagenet}
J.~Deng, W.~Dong, R.~Socher, L.-J. Li, K.~Li, and L.~Fei-Fei, ``Imagenet: A
  large-scale hierarchical image database,'' in \emph{IEEE Conference on
  Computer Vision and Pattern Recognition (CVPR)}, 2009, pp. 248--255.

\bibitem{chen2017rethinking}
L.-C. Chen, G.~Papandreou, F.~Schroff, and H.~Adam, ``Rethinking atrous
  convolution for semantic image segmentation,'' \emph{arXiv preprint
  arXiv:1706.05587}, 2017.

\bibitem{chen2017deeplab}
L.-C. Chen, G.~Papandreou, I.~Kokkinos, K.~Murphy, and A.~L. Yuille, ``Deeplab:
  Semantic image segmentation with deep convolutional nets, atrous convolution,
  and fully connected crfs,'' \emph{IEEE transactions on pattern analysis and
  machine intelligence}, vol.~40, no.~4, pp. 834--848, 2017.

\bibitem{ronneberger2015u}
O.~Ronneberger, P.~Fischer, and T.~Brox, ``U-net: Convolutional networks for
  biomedical image segmentation,'' in \emph{International Conference on Medical
  image computing and computer-assisted intervention (MICCAI)}.\hskip 1em plus
  0.5em minus 0.4em\relax Springer, 2015, pp. 234--241.

\bibitem{newell2016stacked}
A.~Newell, K.~Yang, and J.~Deng, ``Stacked hourglass networks for human pose
  estimation,'' in \emph{Proceedings of the European Conference on Computer
  Vision (ECCV)}.\hskip 1em plus 0.5em minus 0.4em\relax Springer, 2016, pp.
  483--499.

\bibitem{lin2017refinenet}
G.~Lin, A.~Milan, C.~Shen, and I.~Reid, ``Refinenet: Multi-path refinement
  networks for high-resolution semantic segmentation,'' in \emph{Proceedings of
  the IEEE Conference on Computer Vision and Pattern Recognition (CVPR)}, 2017,
  pp. 1925--1934.

\bibitem{fourure2017gridnet}
D.~Fourure, R.~Emonet, E.~Fromont, D.~Muselet, A.~Tr{\'e}meau, and C.~Wolf,
  ``Residual conv-deconv grid network for semantic segmentation,'' in
  \emph{Proceedings of the British Machine Vision Conference (BMVC)}, 2017.

\bibitem{sun2019high}
K.~Sun, Y.~Zhao, B.~Jiang, T.~Cheng, B.~Xiao, D.~Liu, Y.~Mu, X.~Wang, W.~Liu,
  and J.~Wang, ``High-resolution representations for labeling pixels and
  regions,'' \emph{arXiv preprint arXiv:1904.04514}, 2019.

\bibitem{saxena2016convolutional}
S.~Saxena and J.~Verbeek, ``Convolutional neural fabrics,'' in \emph{Advances
  in Neural Information Processing Systems}, 2016, pp. 4053--4061.

\bibitem{andriluka2014}
M.~Andriluka, L.~Pishchulin, P.~Gehler, and B.~Schiele, ``2d human pose
  estimation: New benchmark and state of the art analysis,'' in \emph{IEEE
  Conference on Computer Vision and Pattern Recognition (CVPR)}, June 2014.

\bibitem{mottaghi2014}
R.~Mottaghi, X.~Chen, X.~Liu, N.-G. Cho, S.-W. Lee, S.~Fidler, R.~Urtasun, and
  A.~Yuille, ``The role of context for object detection and semantic
  segmentation in the wild,'' in \emph{IEEE Conference on Computer Vision and
  Pattern Recognition (CVPR)}, 2014.

\bibitem{everingham2015}
M.~Everingham, S.~M.~A. Eslami, L.~Van~Gool, C.~K.~I. Williams, J.~Winn, and
  A.~Zisserman, ``The pascal visual object classes challenge: A
  retrospective,'' \emph{International Journal of Computer Vision}, vol. 111,
  no.~1, pp. 98--136, Jan. 2015.

\bibitem{noh2015learning}
H.~Noh, S.~Hong, and B.~Han, ``Learning deconvolution network for semantic
  segmentation,'' in \emph{Proceedings of the IEEE International Conference on
  Computer Vision}, 2015, pp. 1520--1528.

\bibitem{yang2017learning}
W.~Yang, S.~Li, W.~Ouyang, H.~Li, and X.~Wang, ``Learning feature pyramids for
  human pose estimation,'' in \emph{Proceedings of the IEEE International
  Conference on Computer Vision (ICCV)}, 2017, pp. 1281--1290.

\bibitem{chen2018cascaded}
Y.~Chen, Z.~Wang, Y.~Peng, Z.~Zhang, G.~Yu, and J.~Sun, ``Cascaded pyramid
  network for multi-person pose estimation,'' in \emph{Proceedings of the IEEE
  Conference on Computer Vision and Pattern Recognition (CVPR)}, 2018, pp.
  7103--7112.

\bibitem{rumelhart1988learning}
D.~E. Rumelhart, G.~E. Hinton, R.~J. Williams \emph{et~al.}, ``Learning
  representations by back-propagating errors,'' \emph{Cognitive modeling},
  vol.~5, no.~3, p.~1, 1988.

\bibitem{ioffe2015batch}
S.~Ioffe and C.~Szegedy, ``Batch normalization: Accelerating deep network
  training by reducing internal covariate shift,'' \emph{arXiv preprint
  arXiv:1502.03167}, 2015.

\bibitem{nair2010rectified}
V.~Nair and G.~E. Hinton, ``Rectified linear units improve restricted boltzmann
  machines,'' in \emph{Proceedings of the 27th international conference on
  machine learning (ICML-10)}, 2010, pp. 807--814.

\bibitem{zhang2018shufflenet}
X.~Zhang, X.~Zhou, M.~Lin, and J.~Sun, ``Shufflenet: An extremely efficient
  convolutional neural network for mobile devices,'' in \emph{Proceedings of
  the IEEE Conference on Computer Vision and Pattern Recognition (CVPR)}, 2018,
  pp. 6848--6856.

\bibitem{sun2018igcv3}
K.~Sun, M.~Li, D.~Liu, and J.~Wang, ``Igcv3: Interleaved low-rank group
  convolutions for efficient deep neural networks,'' \emph{arXiv preprint
  arXiv:1806.00178}, 2018.

\bibitem{xie2018interleaved}
G.~Xie, J.~Wang, T.~Zhang, J.~Lai, R.~Hong, and G.-J. Qi, ``Interleaved
  structured sparse convolutional neural networks,'' in \emph{Proceedings of
  the IEEE Conference on Computer Vision and Pattern Recognition}, 2018, pp.
  8847--8856.

\bibitem{zhang2017interleaved}
T.~Zhang, G.-J. Qi, B.~Xiao, and J.~Wang, ``Interleaved group convolutions,''
  in \emph{Proceedings of the IEEE International Conference on Computer
  Vision}, 2017, pp. 4373--4382.

\bibitem{chollet2017xception}
F.~Chollet, ``Xception: Deep learning with depthwise separable convolutions,''
  in \emph{Proceedings of the IEEE Conference on Computer Vision and Pattern
  Recognition (CVPR)}, 2017, pp. 1251--1258.

\bibitem{he2015delving}
K.~He, X.~Zhang, S.~Ren, and J.~Sun, ``Delving deep into rectifiers: Surpassing
  human-level performance on imagenet classification,'' in \emph{Proceedings of
  the IEEE International Conference on Computer Vision (ICCV)}, 2015, pp.
  1026--1034.

\bibitem{howard2017mobilenets}
A.~G. Howard, M.~Zhu, B.~Chen, D.~Kalenichenko, W.~Wang, T.~Weyand,
  M.~Andreetto, and H.~Adam, ``Mobilenets: Efficient convolutional neural
  networks for mobile vision applications,'' \emph{arXiv preprint
  arXiv:1704.04861}, 2017.

\bibitem{sandler2018mobilenetv2}
M.~Sandler, A.~Howard, M.~Zhu, A.~Zhmoginov, and L.-C. Chen, ``Mobilenetv2:
  Inverted residuals and linear bottlenecks,'' in \emph{Proceedings of the IEEE
  Conference on Computer Vision and Pattern Recognition (CVPR)}, 2018, pp.
  4510--4520.

\bibitem{ma2018shufflenet}
N.~Ma, X.~Zhang, H.-T. Zheng, and J.~Sun, ``Shufflenet v2: Practical guidelines
  for efficient cnn architecture design,'' in \emph{Proceedings of the European
  Conference on Computer Vision (ECCV)}, 2018, pp. 116--131.

\bibitem{dumoulin2016guide}
V.~Dumoulin and F.~Visin, ``A guide to convolution arithmetic for deep
  learning,'' \emph{arXiv preprint arXiv:1603.07285}, 2016.

\bibitem{paszke2017automatic}
A.~Paszke, S.~Gross, S.~Chintala, G.~Chanan, E.~Yang, Z.~DeVito, Z.~Lin,
  A.~Desmaison, L.~Antiga, and A.~Lerer, ``Automatic differentiation in
  pytorch,'' 2017.

\bibitem{tian2019decoders}
Z.~Tian, T.~He, C.~Shen, and Y.~Yan, ``Decoders matter for semantic
  segmentation: Data-dependent decoding enables flexible feature aggregation,''
  in \emph{Proceedings of the IEEE Conference on Computer Vision and Pattern
  Recognition}, 2019, pp. 3126--3135.

\bibitem{kingma2014adam}
D.~P. Kingma and J.~Ba, ``Adam: A method for stochastic optimization,''
  \emph{arXiv preprint arXiv:1412.6980}, 2014.

\bibitem{mpii}
\BIBentryALTinterwordspacing
Mpii leader board. [Online]. Available: \url{http://human-pose.mpi-inf.mpg.de.}
\BIBentrySTDinterwordspacing

\bibitem{huang2018condensenet}
G.~Huang, S.~Liu, L.~Van~der Maaten, and K.~Q. Weinberger, ``Condensenet: An
  efficient densenet using learned group convolutions,'' in \emph{Proceedings
  of the IEEE Conference on Computer Vision and Pattern Recognition (CVPR)},
  2018, pp. 2752--2761.

\bibitem{sun2018integral}
X.~Sun, B.~Xiao, F.~Wei, S.~Liang, and Y.~Wei, ``Integral human pose
  regression,'' in \emph{Proceedings of the European Conference on Computer
  Vision (ECCV)}, 2018, pp. 529--545.

\bibitem{lin2014microsoft}
T.-Y. Lin, M.~Maire, S.~Belongie, J.~Hays, P.~Perona, D.~Ramanan,
  P.~Doll{\'a}r, and C.~L. Zitnick, ``Microsoft coco: Common objects in
  context,'' in \emph{Proceedings of the European Conference on Computer Vision
  (ECCV)}.\hskip 1em plus 0.5em minus 0.4em\relax Springer, 2014, pp. 740--755.

\bibitem{andriluka2018posetrack}
M.~Andriluka, U.~Iqbal, E.~Insafutdinov, L.~Pishchulin, A.~Milan, J.~Gall, and
  B.~Schiele, ``Posetrack: A benchmark for human pose estimation and
  tracking,'' in \emph{Proceedings of the IEEE Conference on Computer Vision
  and Pattern Recognition (CVPR)}, 2018, pp. 5167--5176.

\bibitem{insafutdinov2016deepercut}
E.~Insafutdinov, L.~Pishchulin, B.~Andres, M.~Andriluka, and B.~Schiele,
  ``Deepercut: A deeper, stronger, and faster multi-person pose estimation
  model,'' in \emph{European Conference on Computer Vision}.\hskip 1em plus
  0.5em minus 0.4em\relax Springer, 2016, pp. 34--50.

\bibitem{ning2017knowledge}
G.~Ning, Z.~Zhang, and Z.~He, ``Knowledge-guided deep fractal neural networks
  for human pose estimation,'' \emph{IEEE Transactions on Multimedia}, vol.~20,
  no.~5, pp. 1246--1259, 2017.

\bibitem{luvizon2019human}
D.~C. Luvizon, H.~Tabia, and D.~Picard, ``Human pose regression by combining
  indirect part detection and contextual information,'' \emph{Computers \&
  Graphics}, vol.~85, pp. 15--22, 2019.

\bibitem{chu2017multi}
X.~Chu, W.~Yang, W.~Ouyang, C.~Ma, A.~L. Yuille, and X.~Wang, ``Multi-context
  attention for human pose estimation,'' in \emph{Proceedings of the IEEE
  Conference on Computer Vision and Pattern Recognition}, 2017, pp. 1831--1840.

\bibitem{chou2018self}
C.-J. Chou, J.-T. Chien, and H.-T. Chen, ``Self adversarial training for human
  pose estimation,'' in \emph{2018 Asia-Pacific Signal and Information
  Processing Association Annual Summit and Conference (APSIPA ASC)}.\hskip 1em
  plus 0.5em minus 0.4em\relax IEEE, 2018, pp. 17--30.

\bibitem{chen2017adversarial}
Y.~Chen, C.~Shen, X.-S. Wei, L.~Liu, and J.~Yang, ``Adversarial posenet: A
  structure-aware convolutional network for human pose estimation,'' in
  \emph{Proceedings of the IEEE International Conference on Computer Vision},
  2017, pp. 1212--1221.

\bibitem{tang2018deeply}
W.~Tang, P.~Yu, and Y.~Wu, ``Deeply learned compositional models for human pose
  estimation,'' in \emph{Proceedings of the European Conference on Computer
  Vision (ECCV)}, 2018, pp. 190--206.

\bibitem{zhang2018context}
H.~Zhang, K.~Dana, J.~Shi, Z.~Zhang, X.~Wang, A.~Tyagi, and A.~Agrawal,
  ``Context encoding for semantic segmentation,'' in \emph{Proceedings of the
  IEEE Conference on Computer Vision and Pattern Recognition (CVPR)}, 2018, pp.
  7151--7160.

\bibitem{zhou2018unet++}
Z.~Zhou, M.~M.~R. Siddiquee, N.~Tajbakhsh, and J.~Liang, ``Unet++: A nested
  u-net architecture for medical image segmentation,'' in \emph{Deep Learning
  in Medical Image Analysis and Multimodal Learning for Clinical Decision
  Support}.\hskip 1em plus 0.5em minus 0.4em\relax Springer, 2018, pp. 3--11.

\bibitem{zhao2017pyramid}
H.~Zhao, J.~Shi, X.~Qi, X.~Wang, and J.~Jia, ``Pyramid scene parsing network,''
  in \emph{Proceedings of the IEEE Conference on Computer Vision and Pattern
  Recognition (CVPR)}, 2017, pp. 2881--2890.

\bibitem{yuan2018ocnet}
Y.~Yuan and J.~Wang, ``Ocnet: Object context network for scene parsing,''
  \emph{arXiv preprint arXiv:1809.00916}, 2018.

\bibitem{chen2017dual}
Y.~Chen, J.~Li, H.~Xiao, X.~Jin, S.~Yan, and J.~Feng, ``Dual path networks,''
  in \emph{Advances in Neural Information Processing Systems}, 2017, pp.
  4467--4475.

\bibitem{zhu2019deformable}
X.~Zhu, H.~Hu, S.~Lin, and J.~Dai, ``Deformable convnets v2: More deformable,
  better results,'' in \emph{Proceedings of the IEEE Conference on Computer
  Vision and Pattern Recognition (CVPR)}, 2019, pp. 9308--9316.

\bibitem{cordts2016cityscapes}
M.~Cordts, M.~Omran, S.~Ramos, T.~Rehfeld, M.~Enzweiler, R.~Benenson,
  U.~Franke, S.~Roth, and B.~Schiele, ``The cityscapes dataset for semantic
  urban scene understanding,'' in \emph{Proceedings of the IEEE Conference on
  Computer Vision and Pattern Recognition (CVPR)}, 2016, pp. 3213--3223.

\end{thebibliography}
}

\end{document}